\documentclass[journal]{IEEEtran}

\usepackage[utf8]{inputenc}
\usepackage[T1]{fontenc}
\usepackage{amsmath,amssymb,amsfonts}
\usepackage{graphicx}
\usepackage{xcolor}
\usepackage{booktabs}
\usepackage{multirow}
\usepackage{hyperref}
\usepackage{cleveref}
\usepackage{tabularx}
\usepackage{array}
\usepackage{float}
\usepackage{cite}
\usepackage{url}

\newcolumntype{C}[1]{>{\centering\arraybackslash}p{#1}}

\begin{document}

\title{Drop the Hierarchy and Roles:\\How Self-Organizing LLM Agents Outperform Designed Structures}

\author{
\IEEEauthorblockN{Victoria Dochkina}\\
\IEEEauthorblockA{Moscow Institute of Physics and Technology (MIPT),\\
Moscow, Russia\\
dochkina.vs@phystech.edu}
}

\maketitle

\begin{abstract}
We present a 25,000-task computational experiment comparing coordination architectures in multi-agent LLM systems across 8 models, 4--256 agents, and 8 protocols. Our key finding is the \textit{endogeneity paradox}: a hybrid protocol (Sequential) where agent ordering is fixed but role selection is autonomous outperforms both centralized coordination ($+14\%$, $p<0.001$) and fully autonomous protocols ($+44\%$, Cohen's $d=1.86$, $p<0.0001$). Effective self-organization requires both a capable model and the right protocol---neither alone suffices; models below a capability threshold exhibit a reversal where rigid structure outperforms autonomy. The system scales sub-linearly to 256 agents ($p=0.61$) and exhibits emergent properties: dynamic role invention (5,006 unique roles from 8 agents), voluntary self-abstention, and spontaneous hierarchy formation. Results are reproduced across closed-source and open-source models, with open-source achieving 95\% quality at $24\times$ lower cost.
\end{abstract}


\section{Introduction}
\label{sec:introduction}

AI agents need three things to self-organize---and none of them is a pre-assigned role. Given a mission, a communication protocol, and a sufficiently capable model, groups of LLM-based agents spontaneously form organizational structures, invent specialized roles, and voluntarily abstain from tasks outside their competence---outperforming systems with externally designed hierarchies by $14\%$ ($p<0.001$). But remove any of the three ingredients, and the system collapses: without a strong model, self-organization \textit{reverses} and rigid structure becomes necessary; without the right protocol, even the strongest model underperforms.

These are the findings of a 25,000-task computational experiment---the largest to date---comparing coordination architectures in multi-agent systems based on large language models (LLMs). A fundamental question has been overlooked: \textit{what coordination architecture enables the best trade-off between solution quality, cost, scalability, and resilience to disruptions?}

Current research splits into two directions. \textit{Vertical self-improvement} focuses on making individual agents smarter---exemplified by Meta's DGM-Hyperagents~\cite{zhang2026hyperagents}, which achieves open-ended self-improvement through metacognitive self-modification. \textit{Horizontal coordination} addresses how groups of agents collaborate, dominated by systems that replicate human organizational patterns: fixed roles, centralized task allocation, rigid hierarchies~\cite{chen2024chatdev, hong2024metagpt, wu2023autogen, li2024agentverse}. Vertical self-improvement does not answer how multiple agents should coordinate; horizontal frameworks provide structure but may impose unnecessary constraints on agents whose computational nature is fundamentally different from human workers---an LLM agent can instantaneously change specialization, process the full organizational context, and contribute zero marginal cost when idle.

This paper addresses horizontal coordination with a key insight: \textit{effective self-organization requires two conditions simultaneously}---a capable foundation model \textit{and} the right coordination protocol. The protocol unlocks the model's potential, like sheet music unlocks an orchestra; but an orchestra of beginners (weak models) plays better with a conductor than without one.

In this work, we conduct the largest systematic computational experiment on coordination in multi-agent LLM systems to date, spanning:
\begin{itemize}
    \item \textbf{25,000+ task runs} across 20,810 unique configurations;
    \item \textbf{8 LLM models} (closed-source: Claude Sonnet~4.6, GPT-5.4, GPT-4o, GPT-4.1-mini, Gemini-3-flash, GigaChat~2 Max; open-source: DeepSeek v3.2, GLM-5);
    \item \textbf{4 to 256 agents} per system;
    \item \textbf{8 coordination protocols}, from centralized (Coordinator) to fully autonomous (Shared);
    \item \textbf{4 task complexity levels} (L1--L4), from single-domain to adversarial multi-stakeholder tasks.
\end{itemize}

We distinguish between \textit{exogenous} coordination (structure imposed externally) and \textit{endogenous} coordination (structure emerging from within the system). Our central finding is the \textit{endogeneity paradox}: neither maximal external control nor maximal agent autonomy produces optimal results. Instead, a hybrid protocol that provides minimal structural scaffolding (fixed ordering) while allowing maximal role autonomy (self-selected specialization) achieves significantly superior outcomes.

The main contributions of this paper are:

\begin{enumerate}
    \item A framework for characterizing coordination protocols from exogenous (externally controlled) to endogenous (self-organized), with empirical validation across 8 protocols.
    \item The discovery of the endogeneity paradox: the hybrid Sequential protocol outperforms the fully decentralized Shared protocol by $44\%$ in a controlled pilot (Cohen's $d=1.86$, $p<0.0001$), and outperforms the centralized Coordinator by $14\%$ at scale ($p<0.001$).
    \item Evidence that among strong models, coordination protocol choice ($44\%$ quality variation) and model selection ($\sim14\%$) are both critical, with neither alone sufficient for self-organization.
    \item Demonstration of sub-linear scaling from 4 to 256 agents without quality degradation ($p=0.61$), with emergent phenomena including dynamic role invention ($\text{RSI}\to 0$), voluntary self-abstention, and shallow self-organized hierarchies.
    \item Cross-validation of self-organization across closed-source and open-source LLMs, establishing a capability threshold below which self-organization reverses and fixed structure becomes beneficial.
    \item A three-ring constitutional framework for governing autonomous multi-agent organizations.
\end{enumerate}

\section{Related Work}
\label{sec:related}

\subsection{Multi-Agent LLM Systems}

Multi-agent LLM systems have gained significant attention, with several comprehensive surveys mapping the landscape~\cite{guo2024large, xi2023rise, talebirad2023multiagent}. Prominent frameworks include ChatDev~\cite{chen2024chatdev}, which assigns fixed software engineering roles to agents in a waterfall pipeline; MetaGPT~\cite{hong2024metagpt}, which encodes Standard Operating Procedures as inter-agent protocols; and AutoGen~\cite{wu2023autogen}, which provides a conversation-based framework for multi-agent collaboration. AgentVerse~\cite{li2024agentverse} introduces dynamic team formation but retains a centralized ``recruiter'' agent. GPTSwarm~\cite{zhuge2024gptswarm} models agents as optimizable computation graphs, and Mixture-of-Agents~\cite{wang2024mixture} demonstrates that layering LLM outputs improves quality. Recent work on scaling multi-agent collaboration~\cite{chen2024agentverse2} has explored team size effects but with fixed architectures. These systems use \textit{exogenous} coordination: roles, hierarchies, and interaction patterns are designed by humans and fixed before execution.

\subsection{Emergent Coordination and Self-Organization}

Self-organization in multi-agent systems has deep roots in both classical MAS theory~\cite{wooldridge2009multiagent, shoham2009multiagent, dorri2018multiagent} and biological complexity science~\cite{kauffman1993origins, bonabeau1999swarm}. In the LLM era, recent work has explored more autonomous coordination. EvoAgentX~\cite{yuan2025evoagentx} uses evolutionary optimization (TextGrad) to adapt agent populations but requires gradient-based training. AgentNet~\cite{chen2025agentnet} retrieves optimal Directed Acyclic Graphs (DAGs) for agent routing but remains centralized. MAS-ZERO~\cite{li2025maszero} employs a meta-designer for zero-shot multi-agent generation but lacks runtime adaptation. ReSo~\cite{qian2025reso} trains a Contribution Reward Model for DAG optimization, requiring labeled data. HiVA~\cite{wang2025hiva} proposes semantic-topological evolution but has been tested only at small scale.

\subsection{Self-Improving Agent Systems}

A complementary research direction focuses on individual agents that recursively improve themselves. The Darwin G\"{o}del Machine (DGM) and its extension DGM-Hyperagents~\cite{zhang2026hyperagents} achieve impressive open-ended self-improvement through metacognitive self-modification, where the improvement procedure itself is editable. This work advances \textit{vertical} intelligence---making each agent individually stronger. Our work advances \textit{horizontal} intelligence---making groups of agents collectively effective. The two directions are orthogonal and synergistic: stronger individual agents (as produced by Hyperagents-style self-improvement) benefit more from self-organizing coordination protocols (as studied here). Together, they represent two complementary paths toward more capable AI systems.

\subsection{Gap and Positioning}

Existing approaches address different facets of the multi-agent challenge: fixed-architecture coordination~\cite{chen2024chatdev, hong2024metagpt}, training-based adaptation~\cite{yuan2025evoagentx, qian2025reso}, and individual agent self-improvement~\cite{zhang2026hyperagents}. Our study contributes to this landscape by focusing on a question that has received less attention: how does the \textit{degree of agent autonomy in coordination}---from centralized to fully self-organized---affect collective performance at scale? To our knowledge, this is the first work to systematically vary coordination protocols across an exogenous-to-endogenous spectrum, test groups up to 256 agents, compare 8 LLM models, and demonstrate zero-shot runtime self-organization (Table~\ref{tab:positioning}).

\begin{table*}[t]
\centering
\caption{Positioning relative to existing multi-agent LLM systems. $^*$DGM-Hyperagents focuses on single-agent self-improvement (vertical), not multi-agent coordination (horizontal); included for completeness.}
\label{tab:positioning}
\begin{tabular}{lcccccc}
\toprule
\textbf{System} & \textbf{Configs} & \textbf{Max Agents} & \textbf{Models} & \textbf{Protocols} & \textbf{Coordination} & \textbf{Training} \\
\midrule
ChatDev~\cite{chen2024chatdev} & 6 & 8 & 1--2 & 1 & Fixed exogenous & No \\
MetaGPT~\cite{hong2024metagpt} & 4 & 5 & 1 & 1 & Fixed exogenous & No \\
AgentVerse~\cite{li2024agentverse} & 8 & 8 & 1--2 & 3 & Fixed exogenous & No \\
EvoAgentX~\cite{yuan2025evoagentx} & -- & $\sim$10 & 1--2 & 1 & Evolved & Yes (TextGrad) \\
AgentNet~\cite{chen2025agentnet} & -- & $\sim$12 & 1 & 1 & Fixed (DAG) & Yes (retrieval) \\
MAS-ZERO~\cite{li2025maszero} & -- & $\sim$10 & 1 & 1 & Meta-designed & No \\
DGM-Hyperagents~\cite{zhang2026hyperagents} & -- & 1$^*$ & 1 & -- & Self-modifying & Yes (evolutionary) \\
\textbf{This work} & \textbf{20,810} & \textbf{256} & \textbf{8} & \textbf{8} & \textbf{Exo $\to$ endo} & \textbf{No (zero-shot)} \\
\bottomrule
\end{tabular}
\end{table*}

\section{Methodology}
\label{sec:methodology}

\subsection{System Architecture}
\label{sec:architecture}

We model an autonomous AI organization as a discrete-time system. At each time step $t$, a task $\tau_t$ is presented to a group of $N$ LLM agents. The system state $x_t$ encodes agent roles, interaction topology, and accumulated organizational memory. The dynamics follow:

\begin{equation}
    x_{t+1} = F(x_t, u_t, \tau_t, w_t, \varepsilon_t)
\end{equation}

\noindent where $u_t$ represents coordination decisions (protocol, routing), $w_t$ captures external shocks (regulatory changes, agent failures, priority shifts), and $\varepsilon_t$ represents LLM stochasticity.

At each step, we measure five metrics:
\begin{itemize}
    \item $Q_t \in [0.25, 1.0]$: solution quality (multi-criteria LLM-as-judge);
    \item $T_t$: execution time;
    \item $C_t$: cost (token consumption);
    \item $R_t$: risk (compliance failures, errors);
    \item $M_t \in [0.25, 1.0]$: mission relevance.
\end{itemize}

The optimization objective is:
\begin{equation}
    \max_{\{u_t\}} J = \mathbb{E}\left[\sum_{t=1}^{T} \left(\alpha_Q Q_t + \alpha_M M_t - \alpha_T T_t - \alpha_C C_t - \alpha_R R_t\right)\right]
\end{equation}

\noindent with $\alpha_i > 0$ and constraints on acceptable risk and cost levels.

\subsection{Coordination Protocols: From Exogenous to Endogenous}
\label{sec:protocols}

Four primary coordination protocols were evaluated, spanning a spectrum from fully \textit{exogenous} (structure imposed externally) to fully \textit{endogenous} (structure emerging from within):

\textbf{Coordinator (centralized, exogenous):} Agent~0 acts as an external coordinator, analyzing the task and assigning roles and phases to all other agents, who execute in parallel. A single point of control determines the organizational structure. LLM calls: $N+1$ (1 sequential + $N$ parallel).

\textbf{Sequential (hybrid):} Agents process the task in a fixed order. Each agent observes the \textit{completed outputs} of all predecessors and autonomously selects its role, decides whether to participate or abstain, and contributes accordingly. The ordering is exogenous, but role selection and participation decisions are fully \textit{endogenous}. LLM calls: $N$ (sequential). This is analogous to a sports draft where each pick is informed by all previous selections.

\textbf{Broadcast (signal-based, endogenous):} Two rounds---agents first broadcast role intentions simultaneously, then make final decisions informed by all intentions. LLM calls: $2N$ (parallel per round).

\textbf{Shared (fully autonomous, endogenous):} Agents have access to a shared organizational memory (role history from previous tasks) and make all decisions simultaneously and independently. LLM calls: $N$ (parallel).

Four additional bio-inspired protocols (Morphogenetic, Clonal, Stigmergic, Ripple) were also tested; their detailed results will be reported in a forthcoming companion paper.

\subsection{Task Complexity Levels}
\label{sec:tasks}

Tasks span four complexity levels:
\begin{itemize}
    \item \textbf{L1 (Single-domain):} 1 domain, 3--5 steps, no dependencies (e.g., Application Programming Interface (API) security review);
    \item \textbf{L2 (Cross-domain):} 2 domains, 5--10 steps, knowledge integration required;
    \item \textbf{L3 (Multi-phase):} 3+ domains, 10--20 steps, sequential dependencies between phases (e.g., zero-trust migration planning);
    \item \textbf{L4 (Adversarial):} 3+ domains with conflicting stakeholder interests, incomplete information, no single correct answer (e.g., CEO vs.\ Legal vs.\ CFO resource conflicts).
\end{itemize}

\subsection{Evaluation Framework}
\label{sec:evaluation}

Quality assessment uses a multi-criteria \textit{LLM-as-a-judge} methodology. An independent judge model---separate from the agent model to avoid self-evaluation bias---evaluates each solution across five dimensions on a 4-point scale with verbal anchors. The judge model was GPT-4o in Series~1--2 and GPT-5.4 in Series~3; within each series, the judge was held constant across all conditions:

\begin{enumerate}
    \item \textbf{Accuracy} ($s_{\text{acc}}$): factual correctness;
    \item \textbf{Completeness} ($s_{\text{comp}}$): coverage of task requirements;
    \item \textbf{Coherence} ($s_{\text{coh}}$): logical consistency and structure;
    \item \textbf{Actionability} ($s_{\text{act}}$): readiness for practical application;
    \item \textbf{Mission Relevance} ($s_{\text{mis}}$): contribution to organizational mission.
\end{enumerate}

Quality is aggregated as:
\begin{equation}
    Q_t = \frac{s_{\text{acc}} + s_{\text{comp}} + s_{\text{coh}} + s_{\text{act}}}{16}, \quad Q_t \in [0.25, 1.0]
\end{equation}

\noindent The lower bound of $0.25$ arises because each sub-score has a minimum of $1$ (not $0$) on the 4-point scale, so the minimum aggregate is $4/16 = 0.25$.

The evaluation methodology was iteratively refined through four versions (binary $\to$ 10-point $\to$ multi-criteria $5 \times 4$) to achieve stable differentiation. The judge model changed between series (GPT-4o in Series~1--2, GPT-5.4 in Series~3); however, all cross-protocol and cross-model comparisons within each series use the \textit{same} judge, ensuring internal validity. We acknowledge that inter-series comparisons should be interpreted with this caveat in mind.

A Balance Index aggregates all metrics:
\begin{equation}
    B_t = w_Q \hat{Q}_t + w_M \hat{M}_t + w_T \hat{T}_t + w_C \hat{C}_t + w_R \hat{R}_t
\end{equation}
where $\hat{\cdot}$ denotes normalized values and $\sum w_i = 1$ (weights: $w_Q=0.25$, $w_M=0.20$, $w_T=0.20$, $w_C=0.20$, $w_R=0.15$).

\subsection{Experimental Design}
\label{sec:design}

The experiment comprises three series (Table~\ref{tab:experiment_overview}):

\begin{table}[h]
\centering
\caption{Experimental design overview.}
\label{tab:experiment_overview}
\begin{tabular}{clccc}
\toprule
\textbf{Series} & \textbf{Focus} & \textbf{Models} & \textbf{Agents} & \textbf{Tasks} \\
\midrule
1 & Foundational & GPT-4o & 4 & 660 \\
2 & Scaling & GPT-4.1-mini & 4--64 & 8,020 \\
3 & Multi-model & 8 LLMs & 8--256 & 12,130 \\
\midrule
& \textbf{Total} & & & \textbf{$\sim$20,810} \\
\bottomrule
\end{tabular}
\end{table}

\textbf{Series~1 (Foundation, 660 tasks):} Three coordination models were evaluated---Mean-Field Game (fixed roles, population-level coordination), Sparse Graph (topology-dependent coordination across complete, chain, and power-law networks), and Spectral Hierarchy (dynamically emergent hierarchy based on spectral analysis of agent contributions).

\textbf{Series~2 (Scaling, 8,020 tasks):} Nine campaigns tested scaling from 4 to 64 agents across multiple topologies, complexity levels, and shock scenarios using GPT-4.1-mini.

\textbf{Series~3 (Multi-model \& Protocols, 12,130 tasks):} Six blocks evaluated scaling to 256 agents, multi-model comparison (4 models), task complexity (L1--L4), evolutionary hybrids, organizational memory, and self-organizing roles across 4 closed/open-source models with multiple coordination protocols.

\textbf{Shock resilience} was tested by introducing perturbations: random agent removal, hub agent removal, model substitution for 25\% of agents, and regulatory/priority shifts.

\textbf{Reproducibility details.} All API-based models were accessed between February and March 2026 via official APIs or OpenRouter. Temperature was set to $0.7$ for agent models and $0.0$ for judge models to minimize evaluation variance. All prompts, configuration files, and run logs are preserved for reproducibility. We note that API-based models may be updated by providers over time; the exact model snapshots used are identified by run timestamps in our logs.

\section{Results}
\label{sec:results}

\subsection{The Endogeneity Paradox: Protocol Determines Quality}
\label{sec:protocol_results}

\begin{figure}[t]
    \centering
    \includegraphics[width=\columnwidth]{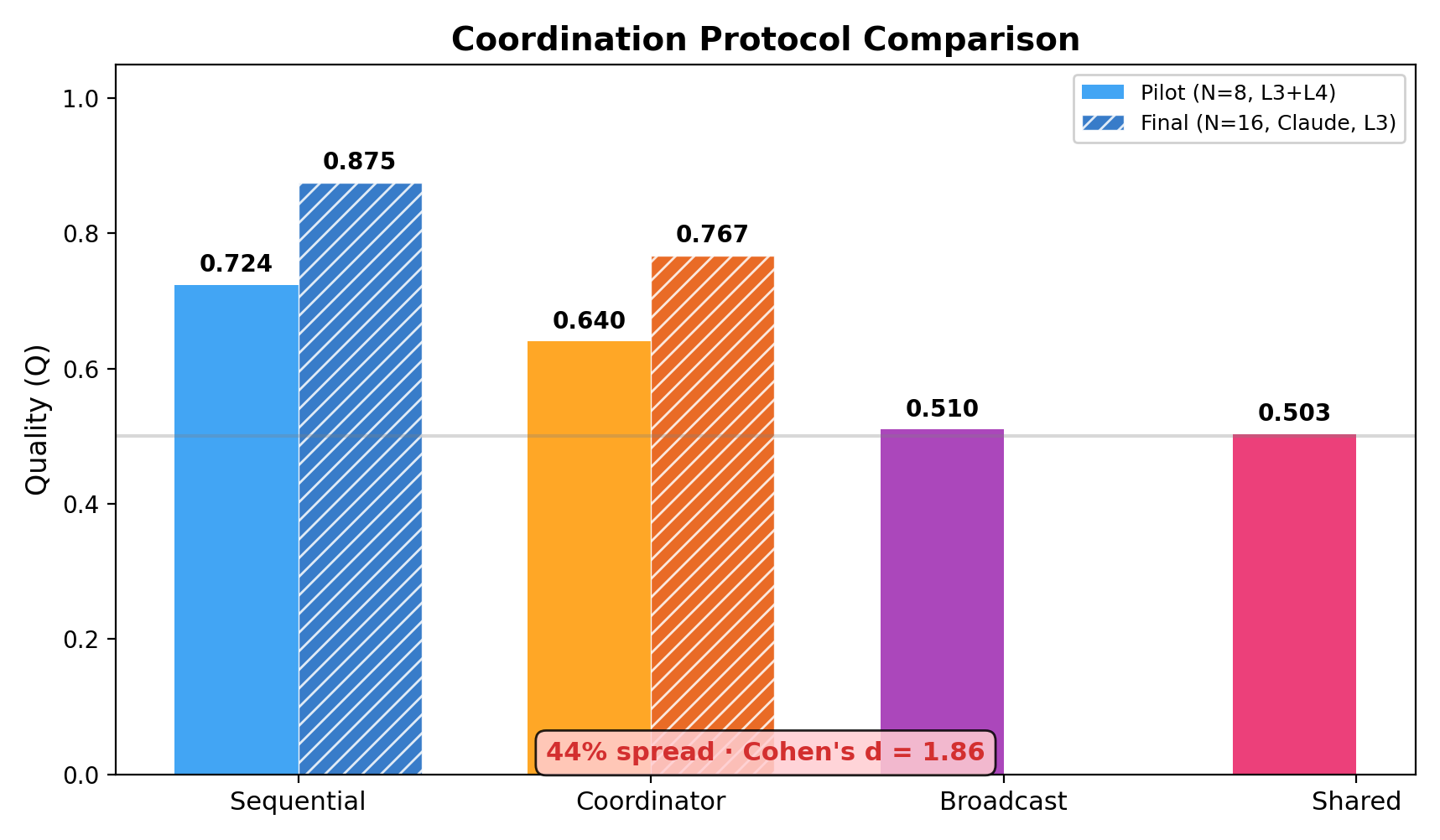}
    \caption{Quality comparison across coordination protocols. Solid bars: pilot ($N=8$, GPT-4.1-mini, L3+L4 average). Hatched bars: final ($N=16$, Claude Sonnet~4.6, L3). The hybrid Sequential protocol achieves the highest quality in both settings.}
    \label{fig:protocols}
\end{figure}

The direct comparison of four protocols under identical conditions (same agents, model, and tasks) revealed that the coordination mechanism has a decisive impact on solution quality---comparable to or exceeding the effect of model choice (Table~\ref{tab:protocols}).

\begin{table}[h]
\centering
\caption{Protocol comparison. Pilot: $N=8$, GPT-4.1-mini, average over L3+L4 tasks. Final ($Q_{\text{L3}}$): $N=16$, Claude Sonnet~4.6, L3 tasks only.}
\label{tab:protocols}
\begin{tabular}{llcccc}
\toprule
\textbf{Protocol} & \textbf{Type} & $Q_{\text{pilot}}$ & $Q_{\text{L3}}$ & \textbf{Balance} & \textbf{Resilience} \\
\midrule
Coordinator & Centralized & 0.640 & 0.767 & 0.478 & 0.774 \\
\textbf{Sequential} & \textbf{Hybrid} & \textbf{0.724} & \textbf{0.875} & \textbf{0.510} & \textbf{0.829} \\
Broadcast & Signal-based & 0.510 & --- & 0.363 & 0.580 \\
Shared & Fully auton. & 0.503 & --- & 0.369 & 0.589 \\
\bottomrule
\end{tabular}
\end{table}

The relative quality difference between the best (Sequential) and worst (Shared) protocol is $44\%$, with effect size Cohen's $d = 1.86$ ($p < 0.0001$).

The paradox lies in the non-monotonic relationship between agent autonomy and quality: neither maximal external control (Coordinator---single point of failure) nor maximal autonomy (Shared---role duplication due to lack of real-time visibility) achieves optimality. The hybrid Sequential protocol succeeds because it provides agents with the optimal information type: not \textit{intentions} (which may change, as in Broadcast), not \textit{history} (which may be stale, as in Shared), not \textit{someone else's plan} (which may be suboptimal, as in Coordinator), but \textit{completed outputs}---the factual results of predecessors in the current task.

\textbf{Cross-model replication.} The Sequential advantage over Coordinator was confirmed across all three tested strong models on L3 tasks (Table~\ref{tab:cross_model}).

\begin{table}[h]
\centering
\caption{Sequential vs.\ Coordinator quality on L3 tasks (N=16, 120 tasks per model, judge: GPT-5.4).}
\label{tab:cross_model}
\begin{tabular}{lcccc}
\toprule
\textbf{Model} & $Q_{\text{Seq}}$ & $Q_{\text{Coord}}$ & $\Delta$ & $p$ \\
\midrule
Claude Sonnet 4.6 & \textbf{0.875} & 0.767 & $+14.1\%$ & $<0.001$ \\
DeepSeek v3.2 & 0.829 & 0.738 & $+12.4\%$ & $<0.001$ \\
GLM-5 & 0.800 & 0.713 & $+12.2\%$ & $<0.001$ \\
\bottomrule
\end{tabular}
\end{table}

\subsection{Scaling: Sub-Linear Cost, Stable Quality}
\label{sec:scaling}

\begin{figure}[t]
    \centering
    \includegraphics[width=\columnwidth]{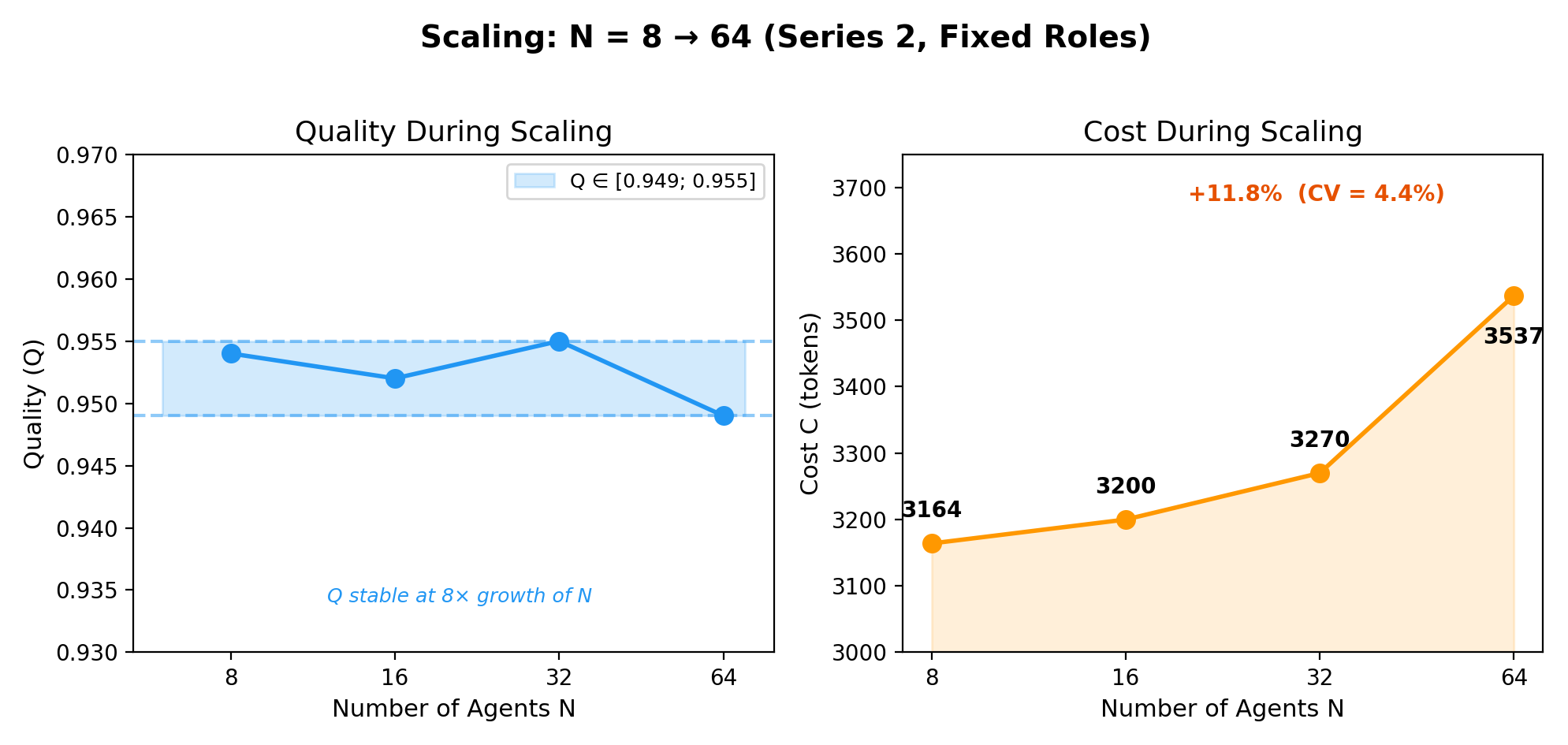}
    \caption{Scaling behavior in Series~2 (fixed roles, GPT-4.1-mini, $N=8 \to 64$). Quality remains stable ($Q \in [0.949, 0.955]$) while cost grows only $11.8\%$ (coefficient of variation CV~$= 4.4\%$) despite an $8\times$ increase in agents.}
    \label{fig:scaling}
\end{figure}

Scaling experiments across two series demonstrated remarkable stability.

\textbf{Series~2 (fixed roles, $N=8 \to 64$, GPT-4.1-mini):} Quality remained virtually constant across an $8\times$ increase in group size (Table~\ref{tab:scaling_s2}), with cost growing only $11.8\%$.

\begin{table}[h]
\centering
\caption{Series~2 scaling (fixed roles, GPT-4.1-mini, 8,020 tasks).}
\label{tab:scaling_s2}
\begin{tabular}{lccc}
\toprule
$N$ & $Q$ & $C$ (tokens) & $T$ (min) \\
\midrule
8 & 0.954 & 3,164 & 14.5 \\
16 & 0.952 & 3,200 & 14.4 \\
32 & 0.955 & 3,270 & 14.0 \\
64 & 0.949 & 3,537 & 14.8 \\
\bottomrule
\end{tabular}
\end{table}

\textbf{Series~3 (self-organization, $N=64 \to 256$, GPT-4.1-mini):} Scaling continued to 256 agents with no statistically significant quality degradation (Table~\ref{tab:scaling_s3}).

\begin{table}[h]
\centering
\caption{Series~3 Block~1 scaling (self-organization, L1 tasks, 6,000 runs).}
\label{tab:scaling_s3}
\begin{tabular}{lccc}
\toprule
$N$ & $Q \pm \sigma$ & Coord.\ overhead & $T$ (min) \\
\midrule
64 & $0.964 \pm 0.039$ & 0.180 & 10.9 \\
96 & $0.962 \pm 0.040$ & 0.180 & 10.9 \\
128 & $0.958 \pm 0.038$ & 0.180 & 11.5 \\
256 & $0.967 \pm 0.032$ & 0.179 & 13.0 \\
\bottomrule
\end{tabular}
\end{table}

The Kruskal-Wallis test yields $H=1.84$, $p=0.61$---\textbf{no statistically significant quality difference} between 64 and 256 agents. Emergent properties intensified with scale: automatic specialization grew from $75\%$ unique roles at $N=4$ to $91\%$ at $N=64$, hierarchy depth increased from 1.0 to 2.0, and adaptation speed improved from 0.7 to 3.0. At $N=256$, approximately $45\%$ of agents became idle through self-abstention, demonstrating an endogenous cost-optimization mechanism.

\subsection{Model Comparison: Quality Ceiling and Capability Threshold}
\label{sec:models}

In Series~3 Block~2, four models were compared under identical topology-based conditions ($N=32$, Exp2 + Exp3 coordination models, 300 tasks each). This setup evaluates raw model capability within structured coordination rather than free-form self-organization (Table~\ref{tab:model_comparison}).

\begin{table}[h]
\centering
\caption{Model comparison in topology-based coordination (N=32, Series~3 Block~2, 300 tasks each). Note: these results reflect structured Exp2/Exp3 topologies; protocol-based results (Section~\ref{sec:open_closed}) show higher quality for strong models under Sequential self-organization.}
\label{tab:model_comparison}
\begin{tabular}{lcccc}
\toprule
\textbf{Model} & $Q$ & $T$ (min) & $C$ (tokens) & $J_{\text{judge}}$ \\
\midrule
DeepSeek v3.2 & \textbf{0.978} & 46.5 & 6,296 & 0.719 \\
GPT-4.1-mini & 0.971 & \textbf{17.5} & 5,059 & \textbf{0.723} \\
Claude Sonnet 4.6 & 0.689 & 36.4 & 5,839 & 0.573 \\
Gemini-3-flash & 0.357 & 45.4 & 4,092 & 0.312 \\
\bottomrule
\end{tabular}
\end{table}

The quality spread between the best (DeepSeek, $Q=0.978$) and worst (Gemini, $Q=0.357$) model is $174\%$. A striking observation is that model rankings shift between experimental setups: Claude Sonnet~4.6 achieves $Q=0.689$ in topology-based experiments (Block~2) but $Q=0.875$ under Sequential self-organization (Block~6, Section~\ref{sec:open_closed}). This difference reflects distinct experimental conditions: Block~2 uses structured Exp2/Exp3 topologies with fixed role assignment ($N=32$), while Block~6 uses free-form self-organization with the Sequential protocol ($N=16$, L3 tasks only). The result is consistent with our central finding: strong reasoning models are disproportionately empowered when given the freedom to self-organize rather than being constrained to pre-designed structures. Among strong models, quality variation attributable to protocol choice ($44\%$) exceeds that of model choice ($\sim14\%$)---but critically, both factors are necessary, as models below the capability threshold show the opposite pattern.

\textbf{Capability threshold for self-organization.} A critical finding is that self-organization is a \textit{privilege of strong models}. When comparing free-form (self-organized) vs.\ fixed-role operation in Series~3 Block~6 ($N=8$, freeform protocol):
\begin{itemize}
    \item \textbf{Claude Sonnet 4.6}: free-form $Q = 0.594$ $>$ fixed $Q = 0.574$ ($+3.5\%$)---autonomy helps;
    \item \textbf{GLM-5}: free-form $Q = 0.519$ $<$ fixed $Q = 0.574$ ($-9.6\%$)---autonomy hurts.
\end{itemize}

Models below the capability threshold show a \textit{reversal effect}: rigid structure helps rather than hinders. The threshold requires three capabilities: (1) \textit{self-reflection}---ability to assess one's own competence (Claude: $8.6\%$ voluntary abstention, GLM-5: $0.8\%$), (2) \textit{deep reasoning}---multi-step logical chains, and (3) \textit{instruction following}---precise adherence to coordination protocols.

\subsection{Closed-Source vs.\ Open-Source Models}
\label{sec:open_closed}

A dedicated experiment (N=16, 120 tasks per model, Sequential and Coordinator protocols) compared Claude Sonnet~4.6 (closed) with DeepSeek v3.2 and GLM-5 (open-source):

\begin{table}[h]
\centering
\caption{Closed vs.\ open-source comparison (N=16, Sequential protocol, judge: GPT-5.4). Token counts and efficiency are for L3 tasks; $Q_{\text{L4}}$ shown for reference.}
\label{tab:open_closed}
\begin{tabular}{lcccc}
\toprule
\textbf{Model} & $Q_{\text{L3}}$ & $Q_{\text{L4}}$ & Tokens & $Q/1\text{K tok}$ \\
\midrule
Claude Sonnet 4.6 & \textbf{0.875} & 0.594 & 37K & \textbf{0.0236} \\
DeepSeek v3.2 & 0.829 & \textbf{0.629} & 47K & 0.0177 \\
GLM-5 & 0.800 & 0.579 & 57K & 0.0140 \\
\bottomrule
\end{tabular}
\end{table}

DeepSeek v3.2 achieves $95\%$ of Claude's L3 quality ($-5.5\%$, $p<0.001$) while being approximately $24\times$ cheaper in API costs. On L4 tasks, DeepSeek shows a trend toward superiority ($+6.0\%$, $p=0.082$). Mission Relevance reaches $4.00/4.00$ for both Claude and DeepSeek on Sequential L3 tasks.

Different models develop distinct self-organization strategies: Claude maximizes role diversity (1,272 unique roles, Gini = 0.055), while DeepSeek with Coordinator employs aggressive agent filtering ($22.4\%$ idle, only 8.5/16 agents active) to achieve maximum cost efficiency ($Q/1\text{K tokens} = 0.026$).

\subsection{Task Complexity and Emergent Hierarchy}
\label{sec:complexity}

\begin{figure}[t]
    \centering
    \includegraphics[width=\columnwidth]{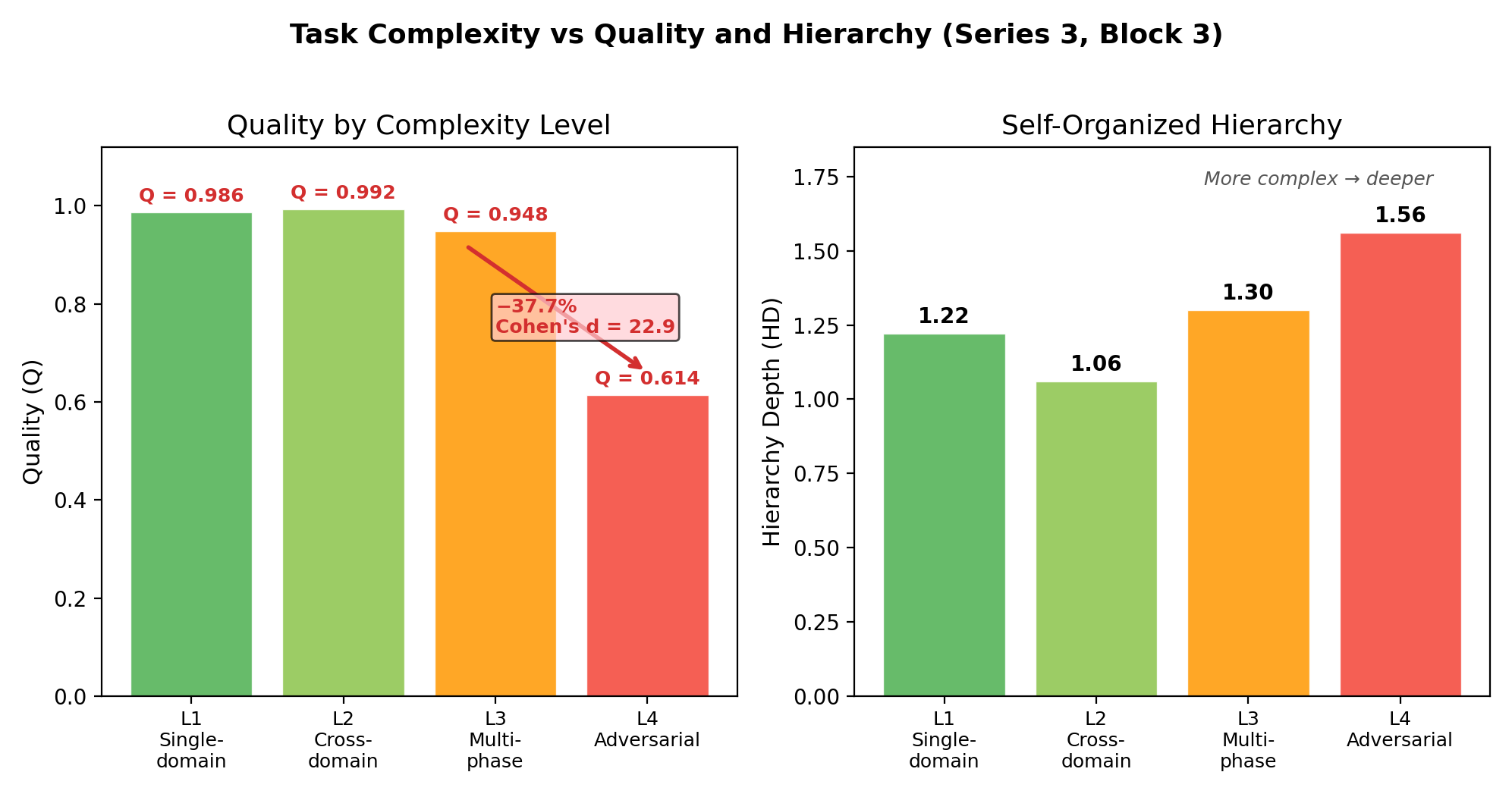}
    \caption{Quality degradation across task complexity levels L1--L4. Hierarchy depth increases with task complexity, indicating emergent structural adaptation.}
    \label{fig:complexity}
\end{figure}

Quality degrades sharply with task complexity, with L4 (adversarial) tasks representing a current frontier (Table~\ref{tab:complexity}). We report the self-organized Hierarchy Depth (HD), measured as the longest chain of agent dependencies in each run.

\begin{table}[h]
\centering
\caption{Quality by task complexity (N=32/64, GPT-4.1-mini, 1,800 tasks).}
\label{tab:complexity}
\begin{tabular}{lcccc}
\toprule
\textbf{Level} & $Q \pm \sigma$ & HD & Completeness & Actionability \\
\midrule
L1 & $0.986 \pm 0.011$ & 1.220 & 3.93 & 3.90 \\
L2 & $0.992 \pm 0.008$ & 1.063 & 3.99 & 3.96 \\
L3 & $0.948 \pm 0.019$ & 1.298 & 3.55 & 3.66 \\
L4 & $0.614 \pm 0.020$ & 1.558 & 1.66 & 2.14 \\
\bottomrule
\end{tabular}
\end{table}

The L1$\to$L4 quality drop of $37.7\%$ has Cohen's $d = 22.9$. Notably, the self-organized Hierarchy Depth (HD) increases from 1.22 (L1) to 1.56 (L4): the system spontaneously develops deeper organizational structures for more complex tasks, without any external instruction to do so.

\subsection{Emergent Properties}
\label{sec:emergent}

\subsubsection{Dynamic Role Specialization}

\begin{figure}[t]
    \centering
    \includegraphics[width=\columnwidth]{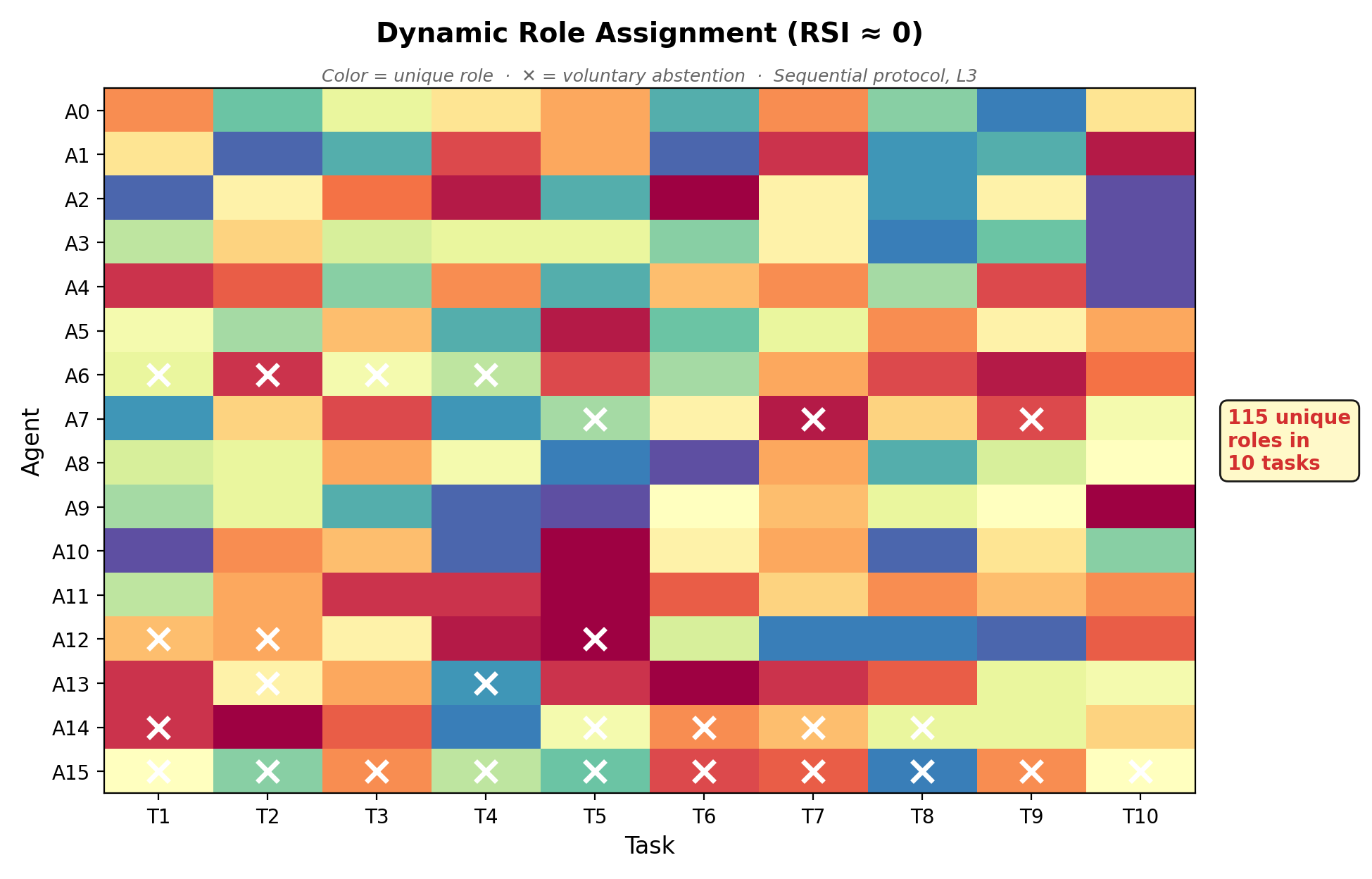}
    \caption{Role assignment heatmap (Sequential protocol, $N=16$, Claude Sonnet~4.6, 10 L3 tasks). Each cell color represents a unique role chosen by the agent for that task; $\times$ marks voluntary abstention. The mosaic pattern (115 unique roles in 10 tasks) demonstrates RSI~$\to 0$: agents reinvent their specialization for each task rather than settling into fixed positions.}
    \label{fig:role_heatmap}
\end{figure}

Across all configurations, RSI converges to zero (Fig.~\ref{fig:role_heatmap}). With 8 agents, 5,006 unique role names were observed; with 64 agents, 5,010 ($+0.1\%$). Agents do not consolidate into fixed positions but reinvent their specialization for each task. At $N=4$, $75\%$ of roles are unique; at $N=64$, $91\%$, with $54\%$ of role names used exactly once.

\subsubsection{Voluntary Self-Abstention}

\begin{figure}[t]
    \centering
    \includegraphics[width=\columnwidth]{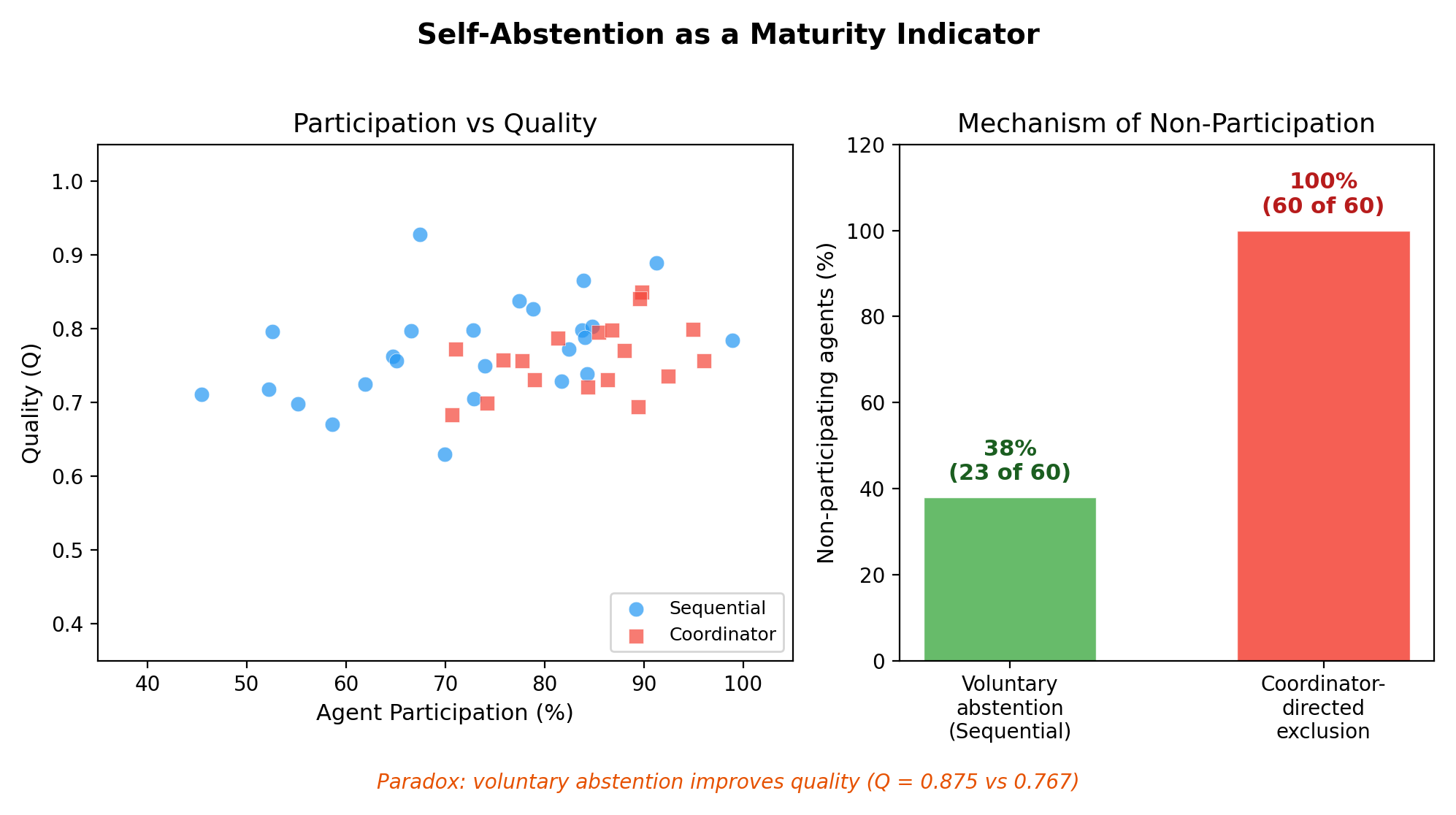}
    \caption{Self-abstention as an emergent property. Left: agent participation rate vs.\ quality across tasks (Sequential = circles, Coordinator = squares). Right: mechanism of non-participation---38\% voluntary (Sequential, endogenous) vs.\ 100\% coordinator-directed (Coordinator, exogenous). Voluntary abstention correlates with higher quality ($Q=0.875$ vs.\ $0.767$).}
    \label{fig:abstention}
\end{figure}

In the Sequential protocol, agents voluntarily abstain from participation when they assess their potential contribution as insufficient (Fig.~\ref{fig:abstention}). In a representative experiment, 38 out of 60 non-contributing agents withdrew by their own decision (endogenous), compared to the Coordinator protocol where all 60 non-contributing agents were excluded by the coordinator's directive (exogenous).

Claude demonstrates the highest voluntary abstention rate ($8.6\%$), while models below the capability threshold show excessive abstention, leading to incomplete task coverage and quality degradation.

\subsubsection{Shallow Self-Organized Hierarchy}
The system consistently prefers flat structures. Hierarchy Depth grows from 1.0 (trivial) to only 2.0 when scaling from 4 to 64 agents, indicating that the system organizes into at most two management layers without external design.

\subsection{Spectral Characteristics of Self-Organization}
\label{sec:spectral}

\begin{figure}[t]
    \centering
    \includegraphics[width=\columnwidth]{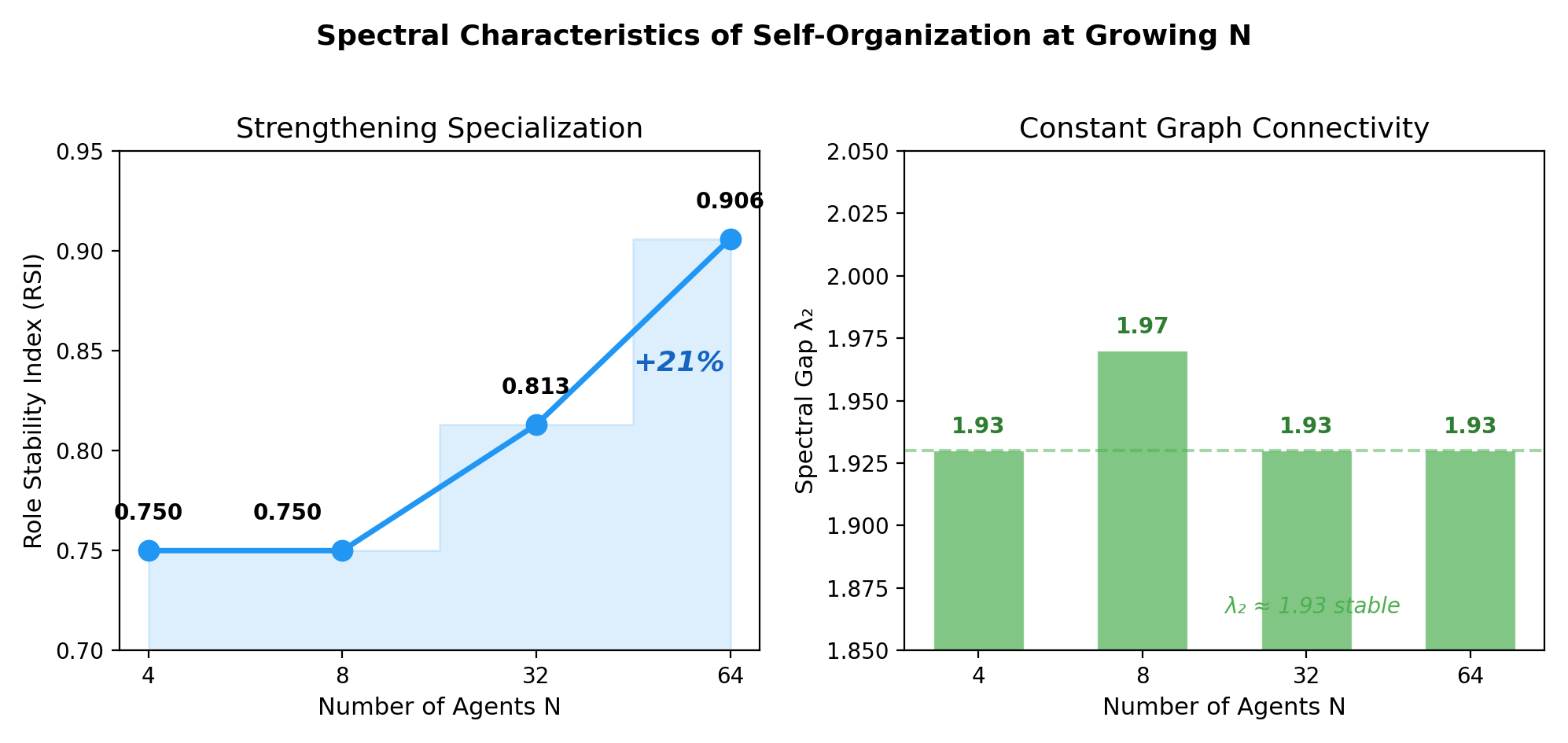}
    \caption{Spectral characteristics of self-organization at growing $N$. Left: Role Stability Index (RSI) increases from 0.750 to 0.906 ($+21\%$), indicating stronger specialization in larger groups. Right: the spectral gap $\lambda_2 \approx 1.93$ remains stable, indicating constant graph connectivity.}
    \label{fig:spectral}
\end{figure}

In the Spectral Hierarchy experiments (Exp3, Series~1--2), where agents possess evolving skill profiles and are assigned to subtasks based on competence, spectral analysis revealed three emergent effects (Fig.~\ref{fig:spectral}):

\begin{enumerate}
    \item \textbf{Strengthening specialization:} The RSI increased from 0.750 ($N=4$) to 0.906 ($N=64$), a $+21\%$ gain---larger groups develop stronger role differentiation when skill profiles accumulate.
    \item \textbf{Flat coordination:} Hierarchy depth remained at 1.0 across all $N$---agents spontaneously form horizontal structures without deepening the hierarchy.
    \item \textbf{Constant connectivity:} The spectral gap $\lambda_2 \approx 1.93$ remained stable across all group sizes, indicating that the interaction graph preserves its connectivity properties during scaling.
\end{enumerate}

This contrasts with the free-form self-organization experiments (Section~\ref{sec:emergent}), where RSI $\to 0$ and agents reinvent their roles from scratch for each task. The difference reflects two regimes: in structured topologies, agents develop persistent specialization; in protocol-based self-organization, maximal role fluidity emerges naturally.

\subsection{Resilience to External Shocks}
\label{sec:resilience}

Self-organizing systems demonstrated rapid recovery from perturbations. Three shock types were tested on $N=32$ agents (Complete topology):
\begin{itemize}
    \item \textbf{Random agent removal:} quality recovers within 1 iteration;
    \item \textbf{Hub agent removal:} quality recovers within 1 iteration;
    \item \textbf{Model substitution (25\% of agents):} quality recovers within 1 iteration.
\end{itemize}

The Spectral Hierarchy model achieved the highest Resilience Index ($\text{RI}=0.959$) with zero quality variance ($\sigma_Q = 0.000$), followed by Sparse Graph ($\text{RI}=0.919$) and Mean-Field ($\text{RI}=0.844$).

Adaptation time ($T_{\text{adapt}}$) improves with system size: $0.7 \to 1.5 \to 3.0$ adaptation speed as $N$ increases, suggesting that larger self-organizing systems heal faster.

\section{Discussion}
\label{sec:discussion}

\subsection{Vertical vs.\ Horizontal Intelligence: Both Are Necessary}

Our results reveal a fundamental interplay in AI system design. Advances in \textit{vertical} intelligence---making individual agents smarter through self-improvement~\cite{zhang2026hyperagents}, scaling, and fine-tuning---and advances in \textit{horizontal} intelligence---designing how groups of agents coordinate---are both essential. A strong model without the right protocol is like an orchestra without sheet music; the right protocol without a capable model is like sheet music without musicians. The two ingredients are multiplicative, not additive: among strong models, protocol choice explains $44\%$ of quality variation (Sequential vs.\ Shared on the same model) while model choice explains $\sim14\%$ (Claude vs.\ GLM-5 on the same protocol). We see these directions as complementary: stronger models amplify the benefits of self-organizing protocols, and better protocols unlock more of each model's potential.

\subsection{The Endogeneity Paradox: Why Neither Control Nor Freedom Wins}

The superiority of the Sequential protocol can be understood through the information each agent receives. Sequential provides \textit{factual, task-specific, accumulated} information: each agent observes what predecessors \textit{actually did} for this particular task. This is informationally superior to:
\begin{itemize}
    \item \textit{Intentions} (Broadcast): may change between rounds;
    \item \textit{Historical patterns} (Shared): may not apply to the current task;
    \item \textit{A single agent's plan} (Coordinator): limited by one agent's judgment.
\end{itemize}

The analogy is to a sports draft: each team selects knowing all previous picks, naturally filling complementary positions without central planning. The paradox is that \textit{minimal structure enables maximal emergence}---one simple constraint (fixed ordering) unlocks spontaneous role differentiation, voluntary abstention, and mission alignment that no amount of explicit design achieves.

\subsection{Role as Emergent Function, Not Pre-Assigned Label}

The near-zero RSI challenges the fundamental assumption underlying most multi-agent LLM architectures---that agents should have fixed roles. For LLM agents, a ``role'' is not a position in an organizational chart but a \textit{computational function activated by task context}. Unlike human organizations, where role switching incurs retraining costs and cognitive overhead, an LLM agent transitions from architect to analyst at zero switching cost. Pre-assigning fixed roles to agents---whether framed as organizational positions or agent role descriptions---is an anti-pattern that replicates human limitations onto entities that lack them.

\subsection{Constitutional Framework for AI Organizations}

Self-organization requires boundaries: the Shared protocol (near-total autonomy) produced the worst results ($Q=0.503$), while Sequential (bounded autonomy) produced the best ($Q=0.724$ pilot; $Q=0.875$ at $N=16$). We propose a \textit{three-ring constitutional framework}:

\textbf{Ring~1 (Immutable Core---human only):} Mission, values, the right to self-abstention. Errors cascade system-wide.

\textbf{Ring~2 (Standards---human + system):} Metrics, governance, audit. The system proposes; humans approve.

\textbf{Ring~3 (Protocols---autonomous):} Coordination parameters, batch sizes, thresholds. Full system autonomy with A/B testing.

The principle: \textit{the closer to ``why,'' the more human control; the closer to ``how,'' the more system autonomy}.

\subsection{Implications: Mission, Protocol, Model---Not Pre-Assigned Roles}

Our findings yield a practical recipe for deploying autonomous multi-agent systems. Instead of assigning roles and writing instructions, practitioners should:

\begin{enumerate}
    \item \textbf{Define mission and values, not role assignments.} An LLM agent is not constrained by a physical body or fixed skill set---pre-assigning it a role replicates human limitations onto entities that lack them. The system achieves Mission Relevance $4.00/4.00$ when given a mission and freedom, not when given a role description.
    \item \textbf{Choose the right protocol.} Among capable models, protocol choice explains $44\%$ of quality variation (Sequential vs.\ Shared, same model, same tasks). The protocol is the amplifier of collective capability.
    \item \textbf{Invest in model quality, not agent quantity.} Scaling from 64 to 256 agents yields no quality improvement ($p=0.61$) at $4.6\times$ cost. The quality spread between models reaches $174\%$---the ``musician'' matters more than the number of seats.
    \item \textbf{Combine models, don't pick one.} DeepSeek achieves $95\%$ of Claude's quality at $24\times$ lower cost. Strong models for L3--L4, efficient models for L1--L2. No single model dominates all dimensions.
\end{enumerate}

\subsection{Limitations}

\begin{enumerate}
    \item \textbf{LLM-as-judge evaluation.} All quality assessments rely on LLM judges (GPT-4o, GPT-5.4) rather than human evaluators. While the multi-criteria decomposition and strict model separation mitigate self-evaluation bias, LLM judges may introduce systematic biases (e.g., preference for verbose responses). A human evaluation study on a representative subset (50--100 tasks) is a priority for future validation.
    \item \textbf{Judge model change.} The judge model changed between series (GPT-4o $\to$ GPT-5.4). All within-series comparisons use the same judge, preserving internal validity, but absolute $Q$ values are not directly comparable across series. Cross-judge calibration on a shared subset is planned as follow-up.
    \item \textbf{Synthetic tasks.} All tasks are synthetically generated to enable controlled comparison. While designed to mirror real-world complexity across four levels (L1--L4), validation on established benchmarks or authentic business workflows would strengthen external validity.
    \item \textbf{Sequential latency.} The Sequential protocol has $O(N)$ latency, which may be prohibitive at very large scale. Batched Sequential variants (Section~\ref{sec:future}) are a promising mitigation.
    \item \textbf{API stability.} API-based experiments depend on provider stability. Some models experienced rate-limiting that reduced task completion rates, introducing potential selection bias.
    \item \textbf{Multiple comparisons.} Multiple statistical tests were performed without formal correction. However, the primary findings (Sequential vs.\ Shared: $p<0.0001$; Sequential vs.\ Coordinator: $p<0.001$ across three models) have $p$-values sufficiently small that they remain significant under Bonferroni correction for the number of comparisons reported ($\alpha_{\text{corrected}} = 0.05/20 = 0.0025$).
\end{enumerate}

\section{Future Work}
\label{sec:future}

Several directions extend this research:

\begin{enumerate}
    \item \textbf{Validation on real-world tasks.} The current experiment uses synthetically generated tasks designed to mirror real-world complexity. A natural next step is deploying Sequential self-organization on authentic business workflows---regulatory analysis, strategic planning, incident response---to validate whether the endogeneity paradox holds in production environments.

    \item \textbf{Batched Sequential for reduced latency.} The Sequential protocol's $O(N)$ latency becomes a bottleneck at large $N$. A batched variant---where groups of $K$ agents work in parallel, then the next group observes all previous outputs---could preserve the informational advantage of Sequential while achieving $O(N/K)$ latency. This represents a promising middle ground between Sequential's quality and Broadcast's speed.

    \item \textbf{Bio-inspired coordination protocols.} Four additional protocols inspired by biological coordination mechanisms (Morphogenetic, Clonal, Stigmergic, Ripple) have been tested in preliminary experiments and will be reported in a forthcoming companion paper. Early results suggest that Ripple (wave-based information propagation) achieves quality comparable to Sequential on adversarial tasks while permitting greater parallelism.

    \item \textbf{Constitutional governance in practice.} The three-ring constitutional framework proposed in this paper is currently a theoretical model. Testing it as a live governance mechanism---where Ring~3 protocols are autonomously optimized by the system via A/B testing while Ring~1 mission constraints remain human-controlled---would validate the framework's practical viability.

    \item \textbf{Combining vertical and horizontal advances.} Our results and recent work on self-improving agents~\cite{zhang2026hyperagents} suggest a multiplicative relationship between model capability and coordination protocol. A compelling next experiment would deploy self-improving agents \textit{within} a Sequential coordination framework to test whether vertical and horizontal intelligence gains compound.
\end{enumerate}

\section{Conclusion}
\label{sec:conclusion}

This paper presents the largest systematic study of coordination in multi-agent LLM systems, spanning over 25,000 task runs across 8 models, 4--256 agents, 8 protocols, and 4 complexity levels. Our central contribution is the discovery of the \textit{endogeneity paradox}: optimal coordination emerges not from maximal control or maximal autonomy, but from a hybrid that provides minimal structural scaffolding (fixed ordering) while enabling full role autonomy (endogenous specialization). In a controlled comparison, the Sequential protocol outperforms fully autonomous coordination by $44\%$ (Cohen's $d = 1.86$, $p < 0.0001$); at scale ($N=16$, L3), it reaches $Q=0.875$, outperforming centralized coordination by $14\%$ ($p<0.001$).

We demonstrate that both protocol choice and model capability are critical design decisions: among strong models, protocol variation accounts for $44\%$ of quality differences while model variation accounts for $\sim14\%$; but neither alone suffices, and scaling team size beyond 64 agents yields no improvement ($p=0.61$). The system exhibits emergent properties---dynamic role invention (RSI $\to 0$), voluntary self-abstention, spectral stability ($\lambda_2 \approx 1.93$), and spontaneous hierarchy formation---that are reproduced across both closed-source and open-source models. Open-source models achieve $95\%$ of closed-source quality at $24\times$ lower cost, validating the viability of multi-model strategies.

The practical recipe is simple: \textit{give agents a mission, a protocol, and a capable model---not a pre-assigned role.} Self-organization will do the rest.

\section*{Acknowledgments}
The author thanks S.~Budyonny for scientific supervision, and E.~Latypova and K.~Ruppel for methodological and organizational support during the experimental phase.
This article includes text that was refined with the assistance of an AI language model (Claude, Anthropic). The AI tool was used for language editing, formatting, and structuring of the manuscript text. All scientific content, experimental design, data analysis, and intellectual contributions are solely the work of the author.

\section*{Data Availability}
The experimental data, code, analysis scripts, and all LLM prompts used in this study will be made publicly available in a dedicated repository upon acceptance. Run logs with timestamps, model identifiers, and raw judge evaluations are preserved for full reproducibility.

\bibliographystyle{IEEEtran}

\end{document}